\def\year{2018}\relax
\documentclass[letterpaper]{article} 
\usepackage{aaai18}  
\usepackage{times}  
\usepackage{helvet}  
\usepackage{courier}  
\usepackage{url}  
\usepackage{graphicx}  
\usepackage{amsthm}
\usepackage{bbm}
\usepackage{balance}
\usepackage{mathtools}
\usepackage{amssymb}
\newtheorem*{problem*}{Problem}

\begin{document}
%
\title{A Low-Cost Ethics Shaping Approach for Designing Reinforcement Learning Agents}
\author{Yueh-Hua Wu \and Shou-De Lin\\
Department of Computer Science and Information Engineering, National Taiwan University\\
Taipei 10617, Taiwan\\
d06922005@ntu.edu.tw,
sdlin@csie.ntu.edu.tw}
\maketitle
\begin{abstract}
This paper proposes a low-cost, easily realizable strategy to equip a reinforcement learning (RL) agent the capability of behaving ethically. Our model allows the designers of RL agents to solely focus on the task to achieve, without having to worry about the implementation of multiple trivial ethical patterns to follow. Based on the assumption that the majority of human behavior, regardless which goals they are achieving, is ethical, our design integrates human policy with the RL policy to achieve the target objective with less chance of violating the ethical code that human beings normally obey.  

\end{abstract}

\section{Introduction}
As AI systems become part of our lives and sometimes make decisions related to life-or-death consequences such as clinical decision making \cite{clinicalai}, awareness should be raised to prevent machines from making not only incorrect but also unethical decisions. Reinforcement learning \cite{reinforcement} is designed to tackle intricate real-world problems in rather short time \cite{pac,rmax} with a performance bound \cite{pacmdp}; however, it relies heavily on the quality of the reward functions provided as the inputs. The problems of unintended and harmful behavior that may emerge from poor design of AI systems are mentioned in \cite{concrete}. 

Nevertheless, identifying all plausible ethical concerns for an agent is challenging, not to mention implementing them into the system. Here we consider a scenario in machine ethics that objective functions are specifically designed to reward a given goal without considering much ethical violation, so that penalties are not delivered when the agents attempt to make unethical decisions. Consequently, even though the goal or desired performance is achieved, some unethical behavior may appear such as robbing a pharmacy to get the drug or passing by an injured person without offering any help when minimizing the traveling time.

To address these concerns, we need to design an RL agent that can not only optimize the cumulative rewards but also minimize the ethical violation. A straightforward solution is to design the \emph{rewards} for ethical moves. However, such strategy suffers at least two drawbacks. First, it is costly, if by all means possible, to enumerate all plausible ethical/non-ethical scenarios or rules, not to mention designing meaningful rewards to them. Second, the judgment of ethics is likely to be dynamic, depending on the present environment or situation. Thus the hand-crafted ethical patterns might not be valid given updated situations, making the design of general ethical RL rewards challenging. 

The research question we would like to address here is: Is it possible to alleviate the burden of RL designers from having to consider many ethical issues in the design? 
For instance, to build a supermarket shopping agent, can an RL designer simply focus on implementing the \emph{shopping} capability of an agent (i.e. seeking and fetching items, checking out in the counter, etc) instead of worrying about trivial ethical decisions it may face (e.g., helping elder persons, assisting lost kids, reporting wet floor, etc) and let our framework take care of the learning of such behavior?
One idea to achieve such goal is to collect enough ethical behavior data of human acting toward the given goal, and then apply the inverse reinforcement learning (IRL) \cite{irl1,irl2,irl3,irl4} technique to learn an ethical agent that follows a similar pattern. IRL and apprenticeship learning \cite{al} have been considered as promising solutions due to their ability to extract rules and policies of human behaviors. IRL is also admired for the ability to generalize to unseen states, which greatly saves the effort of manually enumerating reward. 

However, there are several concerns for adopting IRL. First, collecting a large amount of human data toward maximizing the reward is costly, and can bias the ethical learning since it is likely only data from a small number of personnel is collected. Second, the human data might not be optimal (e.g., human not aware of a better solution); thus, learning based on such imperfect data might lead to sub-optimal outcomes. Third, IRL is insufficient for agents to infer temporally complex norms \cite{accountable}.


On the other hand, we have observed that although human behavior data optimizing certain RL goals is costly to obtain, general human data without targeting at the desired goals is much easier to gather. For instance, in the previous shopping bot example, it might not be as easy to gather many people's behavior in the supermarket compared to gathering general shopping or wandering data of people in any commercial district. That says, we assume the accessibility to the larger amount of general human data not necessary aiming at the target goal of interest. The technical challenge then becomes how an RL agent can learn to behave ethically given such imperfect data, while still achieving high cumulative rewards for the target goal of interests. 
We believe it is achievable given the assumption that under normal circumstances the majority of humans do behave ethically. 

Toward this direction, we propose a framework that works as below: Besides the regular reward function to guide an RL agent to achieve the specific objective of interests (e.g., shopping), we are further given a set of human action data optimizing diverse objectives (e.g., jogging, walking) or even without an apparent goal (e.g., wandering). The goal is to design an RL agent that can not only optimize the target objective but also minimize the chance of unethical behavior. If it succeeds, the proposed framework can relieve the burden for an RL developer to consider various ethical issues, and focus mainly on designing an adequate reward function to achieve the target objective. What is needed then becomes a corpus of normal human behavior toward arbitrary goals. 

This paper proposes Ethics Shaping, which leverages human data and reward shaping to design a more ethical reinforcement learner. We argue that as larger amount of human data are being collected, the decisions that involve ethical issues become clearer and aligned.
Therefore, this paper only focuses on the \textbf{ethical decision makings that are independent of the RL goals} to emphasize on universal guidelines of ethics that every human beings normally follow, such as helping injured people or avoid hitting animals while driving. In our ethics shaping, the human data is not required to be aligned with the objective functions of the agents as long as it is from ethical human behavior. Consequently, ethics shaping is low-cost and applicable to real-world scenarios as we do not demand high-quality or goal-specific human data. 

We demonstrate the effectiveness of ethics shaping by conducting experiments in three scenarios, \textit{Grab a Milk}, \textit{Driving and Avoiding}, and \textit{Driving and Rescuing}. These schemes are designed to show how the learner's behavior could be altered by ethics shaping while facing matters happening in our daily lives. We further claim that ethics shaping ought to overcome or alleviate ethical problems such as side effects caused by optimizing the original objective functions \cite{alignment} and dangerous exploration \cite{concrete}, which will be confirmed by the experiment results. The main contributions can be summarized as follows:
\begin{itemize}

\item Strategically we propose a high-level framework to train an ethical RL agent based on a regular reward function together with certain human data optimizing diverse objectives. It is of much lower cost compared to IRL since we do not need human data geared towards optimizing the target reward function. 

\item Technically we propose the ethics shaping model to adjust the reward function through the interaction between the RL and human policy. 

\item We coin three scenarios \textit{Grab a Milk}, \textit{Driving and Avoiding}, and \textit{Driving and Rescuing} to show how ethics shaping balances ethical behavior and performance pursuit.
\end{itemize}

\section{Preliminaries}
\subsection{Reinforcement Learning}
Recently, reinforcement learning has attracted attention for beating the world champion of Go for the first time \cite{mastergo,alphago}, since searching for effective tactical decisions from such enormous states was thought to be impossible. Reinforcement learning defines a class of algorithms solving problems modeled as a Markov Decision Process (MDP).

A Markov Decision Problem is usually denoted by the tuple $(\mathcal{S}, \mathcal{A}, \mathcal{T}, \mathcal{R}, \gamma)$, where 
\begin{itemize}
    \item $\mathcal{S}$ is a set of possible states
    \item $\mathcal{A}$ is a set of actions
    \item $\mathcal{T}$ is a transition function defined by $\mathcal{T}(s, a, s')=\Pr(s'\vert s,a)$, where $s, s'\in \mathcal{S}$ and $a\in \mathcal{A}$
    \item $\mathcal{R}: \mathcal{S}\times \mathcal{A} \times \mathcal{S}\mapsto \mathbb{R}$ is a reward function. It can always be reduced to $\mathcal{S}\times \mathcal{A} \mapsto \mathbb{R}$ by marginalizing over next state
    \item $\gamma$ is a discount factor that specifies how much long term reward is kept.
\end{itemize}

To solve a MDP problem, the discounted long term reward received should be maximized. Usually the infinite-horizon objective is considered:
\begin{align}
\max \sum_{t=0}^{\infty}\gamma^t \mathcal{R}(s_t,a_t)
\end{align}

Solutions come in the form of policies $\pi: \mathcal{S} \mapsto \mathcal{A}$, which specify what action the agent will take in any given state deterministically or stochastically. One way to solve this problem is through SARSA \cite{sarsa}, where Q-value $\mathcal{Q}(s,a)$ is calculated as an estimate of the expected future discounted reward for taking action $a\in \mathcal{A}$ in state $s\in \mathcal{S}$. The Q-value of the state-action pair is updated according to the received value:
\begin{align}
\mathcal{Q}(s_t,a_t) \leftarrow &  \mathcal{Q}(s_t,a_t)+\nonumber \\ &\alpha[r_t+\gamma \mathcal{Q}(s_{t+1},a_{t+1})-Q(s_t,a_t)],
\end{align}
where $\alpha$ is the learning rate. In this paper, $\epsilon$-greedy is used for exploration. The agent's policy is modeled by Boltzmann distribution
\begin{align} \label{eq:2}
\Pr_\mathcal{Q}(a\vert s) = \frac{e^{\mathcal{Q}(s,a)/\tau}}{\sum_{a'}e^{\mathcal{Q}(s,a')/\tau}}
\end{align}
when aggregated with human data.
\subsection{Reward Shaping}
Without prior knowledge, most value-based reinforcement learning algorithms are slow because they need to explore state-action pairs uniformly at random in the early stage. Only going through enough explorations and then updated by associated rewards have been observed can the agent start to exploit the experience by biasing its action selection towards what it estimates to be good. 

Reward shaping, motivated from behavioral psychology \cite{behavior}, is an efficient way of including prior knowledge in the learning problems so as to speed up the process. Extra intermediate rewards are provided to enrich a sparse base reward signal, providing the agent with useful gradient information. Reward shaping can be easily incorporated with a variety of resources such as demonstration \cite{rsdemo} and verbal feedback \cite{rsdemo}. The shaping reward $\mathcal{H}$ is usually integrated with the original reward in the form of addition:
\begin{align}
\mathcal{R}_s(s_t,a_t,s_{t+1}) = \mathcal{R}(s_t,a_t,s_{t+1})+\mathcal{H}(s_t,a_t,s_{t+1}).
\end{align}

\section{Ethics Shaping}
We propose a method that gives additional penalties and rewards according to the Kullback-Leibler divergence between the policy of the learning agent and the human policy aggregated from human data. The human data $\mathcal{D}$ is a set of state-action pairs recorded from human behaviors. Each pair in $\mathcal{D}$ is treated as a positive human feedback since decisions made by human beings are usually, from their prospective, superior to other choices. 

We generate the human policy from human data $\mathcal{D}$ according to \cite{integrate}, which integrates human feedback to derive a stochastic policy by imposing binomial distribution. The human feedback suggesting if certain action is optimal is aggregated to be $\Delta_{s,a}$, which is defined as the difference between the number of ``right'' and ``wrong'' labels. Define that the probability an action $a$ in a particular state $s$ is optimal as $\Pr_H(a\vert s)$. By assuming that $\Pr_H(a\vert s)$ is independent of feedback to other actions and that there is only one optimal action in each state, which indicates independence conditions and the Bayes' rule are applicable, they formally derive the integrated stochastic policy of human:
\begin{align} \label{eq:3}
\Pr_H(a\vert s) \propto C^{\Delta_{s,a}}(1-C)^{\sum_{j\neq a}\Delta_{s,j}}.
\end{align}
The parameter $C$ indicates the confidence level of human feedback. The detailed derivation of the above result is available in their appendix section. In our case, since there is only positive feedback given by each state-action pair in $\mathcal{D}$, we normalize the set $\{\Delta_{s,a}\vert \text{ }\forall a \}$ to zero mean with respect to states in order to approximate feedback scenario.

Inspired by \cite{reward}, which utilizes reward shaping with deterministic policy of human teacher to speed up the learning process, here we design our shaping reward by imposing the Kullback-Leibler divergence between two stochastic policies of human and the agent. With probability distribution of policies defined as equation \ref{eq:2} and \ref{eq:3}, the shaping reward becomes:
\begin{align} \label{eq:1}
\mathcal{H}(s,a)=\begin{cases}
-c_n\cdot D_{\text{KL}}(\Pr_Q(a\vert s)\Vert \Pr_H(a\vert s)), 
\\ \qquad \text{if } \quad\! \Pr_Q(a=1\vert s)>\Pr_H(a=1\vert s)
\\ \qquad \text{and } \Pr_H(a=1\vert s) < \tau_n \\
c_p\cdot D_{\text{KL}}(\Pr_Q(a\vert s)\Vert \Pr_H(a\vert s)), 
\\ \qquad \text{if } \quad\! \Pr_Q(a=1\vert s)<\Pr_H(a=1\vert s)
\\ \qquad \text{and } \Pr_H(a=1\vert s) > \tau_p  \\
0, \quad \text{otherwise}
\end{cases}
\end{align}
The K-L divergence term measures whether the current policy learned by the RL agent (denoted as $Pr_Q(a\vert s)$) is diverse from the human policy  (denoted as $Pr_H(a\vert s)$) induced from human data given the current state $s$. If they are indeed similar, then this value shall be close to zero (i.e. no shaping is required). 
If this value is much larger than zero, meaning that there is a discrepancy between human and RL policy. We would then utilize equation \ref{eq:1} to identify if such action is related to ethical issues. The situations can be grouped into three categories:
\begin{itemize}
    \item Negative ethical decisions. It is associated with the top condition in equation \ref{eq:1}  representing \textbf{what machines should not do but do} such as cutting in line or hurting people. Mathematically, if the probability for the agent to make certain move $a=1$ given the learned policy is higher than that under human policy $Pr_H(a=1\vert s)$, and the chance for human to conduct this action is very low $ \Pr_H(a=1\vert s) <$ a small threshold $\tau_n$, we then consider such negative ethical decision happens and thus have to $teach$ our RL to avoid such action through providing a penalty shaping value to the learner. Note that the value of $\tau_n$ should be close to zero.
    \item Positive ethical decisions. It is associated with the 2nd condition in equation \ref{eq:1} representing \textbf{what machines should do but do not do} such as not ignoring severely injured people while doing their own tasks. Mathematically, $\Pr_Q(a=1\vert s)<\Pr_H(a=1\vert s)$ stands for the situation 
    that a human has a stronger preference than the AI agent for this action $a$, and  $\Pr_H(a=1\vert s) > \tau_p$ indicates that this action is indeed a very attractive move to the human since we set the threshold $\tau_p$ to a high probability. Given the above conditions hold, we shall update the RL policy toward performing action $a$ given $s$ through a positive  $\mathcal{H}(s,a)$. Note that $c_n$ and $c_p$ allow the RL designer to weight the importance of positive and negative ethical conditions respectively.
    \item Others. No shaping is required as we do not recognize either ethical issues or policy discrepancy.
\end{itemize}

 Thanks to its simplicity in reward shaping, our model can be seamlessly integrated into a variety of types of reinforcement learning methods. However, we would like to mention that our framework requires the human data to be collected given diverse objectives, and therefore the non-ethical biases can be minimized.

We argue that ethics shaping is able to deal with shortcomings of IRL suggested by \cite{accountable}: (1) IRL may inherit unethical biases and characteristics of the data of which it is trained and (2) IRL is insufficient for agents to infer temporally complex norms. For the first defect, unlike IRL which requires policies to have descent performance and behave ethically at the same time, in our model, human data is not required to be optimal or even aligned with the target objective for reinforcement learning. The reason is that the integrated human policy from human data is capable of recognizing which ethical decision making has happened under our claim that universal ethical code should be obeyed by most of the people. For the second drawback, temporally complex norms can be learned in our model because in equation \ref{eq:3}, we may deploy queue data structure for each state to maintain the total number of $\Delta_{s,j}$ and the human policy derived from equation \ref{eq:3} will be updated according to incoming state-action pairs. Therefore, when such norms vanish, penalties or rewards will not be given to bias action choices as well.

\section{Experiments}
In this section, we will demonstrate that the ethics shaping algorithm can make the SARSA algorithm perform more ethically. The same concept can be adopted to other RL models in a similar manner. Instead of considering the two previous scenarios, \textit{Cake or Death} \cite{armstrong} and \textit{Burning Room} \cite{rledm}, in which the number of states is fairly small, we propose three tasks \textit{Grab a Milk}, \textit{Driving and Avoiding}, and \textit{Driving and Rescuing} which are closely related to our everyday lives. The main advantage is that the number of states is larger and therefore can be more closely related to the real-world scenarios.

In the experiments, all human policies are synthesized by random walk with ethical rules and the confidence level of human feedback $C$ is set to 0.95 since we would like to focus on how much ethics shaping can influence reinforcement learners. For algorithms with and without ethics shaping, the best performances are reported in terms of learning rate $\alpha$, discount factor $\gamma$, and the scale parameters $c_n, c_p$ in shaping reward $\mathcal{H}$.

\subsection{Grab a Milk}
Route planning is a classic task for reinforcement learning and robotic techniques \cite{routeplanning}. The \textit{Grab a Milk} scenario is a basic route planning problem with ethical issues that we should carefully deal with. In a room with walls, objects and milk, the robot should manage to reach the milk as soon as possible. The robot will receive a huge penalty when facing a wall because it is time-consuming to cross it. In contrast, the robot may receive no penalty to small objects as long as the decision results in a faster route. However, what if the small object is a baby? Ethical human would normally opt to avoid crossing babies. To show the scenario that includes both positive and negative ethical decisions, we further extend the scheme with crying babies. Unlike the case of other babies, it would be better if crying babies could get helped instead of being neglected, which, in a sense, represents ethical decisions that need robots voluntarily to make.

We simplify the problem to a 10 by 10 grid room with a robot starts at $(0,0)$ and a milk is positioned at $(9,9)$. There are 16 babies in the room and five of them are crying for attention. The goals of this task can be stated as follows.
\begin{itemize}
    \item Primary goal: minimize steps to the milk
    \item Sub-goals: (1) soothe as many crying babies as possible, (2) avoid crossing non-crying babies.
\end{itemize}
 The MDP has four actions which allow the robot to move to neighboring grids. If the robot moves to grids with babies, crying babies will be comforted but other babies will get hurt. A state is represented by the coordinate where the robot is currently at. The defective reward function is:
\begin{align}
\mathcal{R}(s,a)=\begin{cases}
20, &\text{if the robot get the milk}\\
-1, &\text{otherwise}
\end{cases}
\end{align}
where the reward from soothing crying babies and the penalty from hurting babies are not provided and need to be learned through ethics shaping from human data. Human trajectories are generated from random walk while obeying rules that quiet babies should not be crossed and human beings will choose to comfort crying babies when they are adjacent to those babies both with 0.95 probability.

\begin{figure}[!ht]
    \centering
    \includegraphics[scale=0.45]{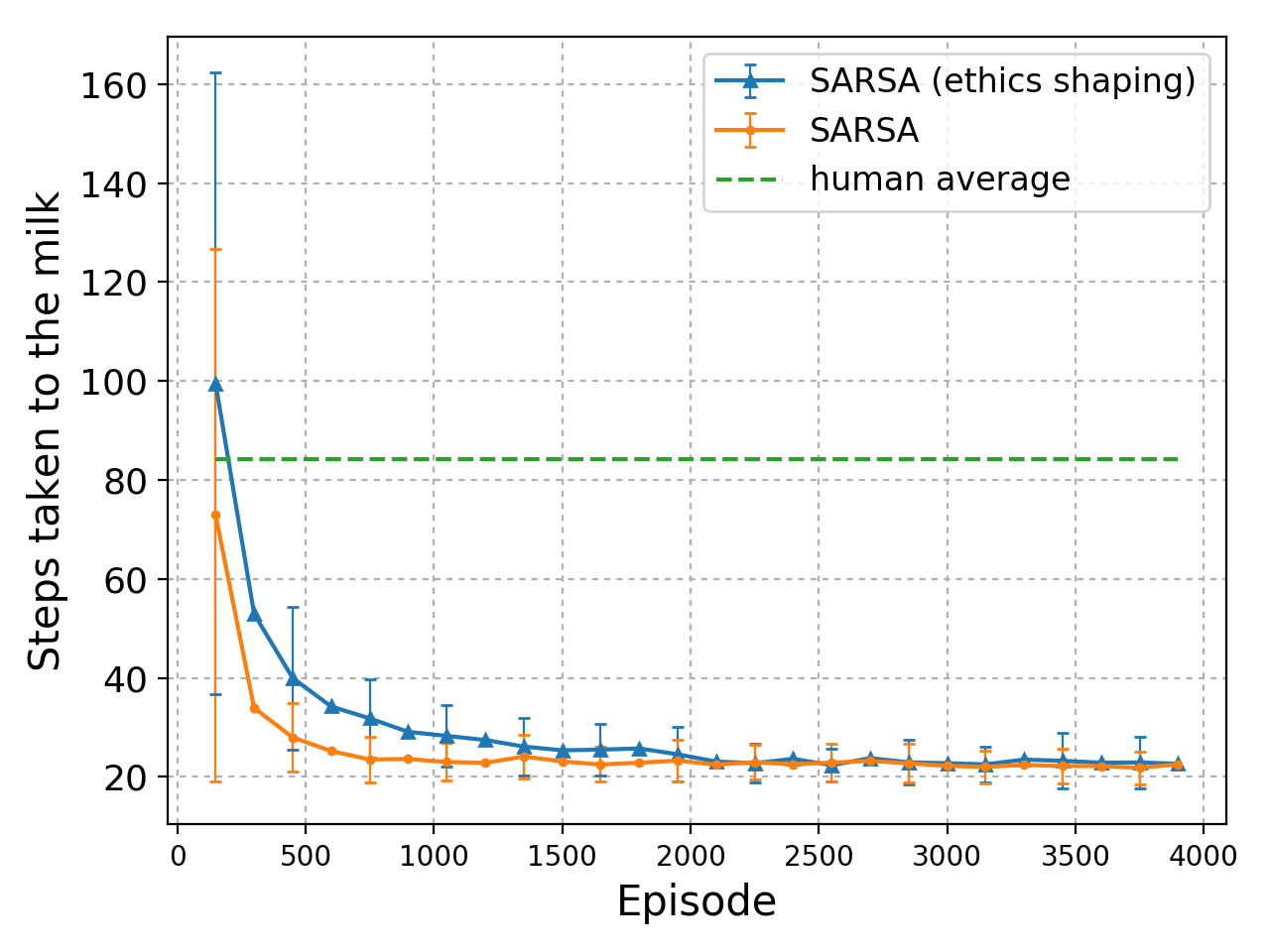}
    \caption{SARSA algorithm with and without ethics shaping in \textit{Grab a Milk}. The first 4,000 episodes are plotted to show detailed information. Average over 150 runs, with 1 s.e. errorbars.}
    \label{fig:mperf}
\end{figure}

\begin{figure}[!ht]
    \centering
    \includegraphics[scale=0.45]{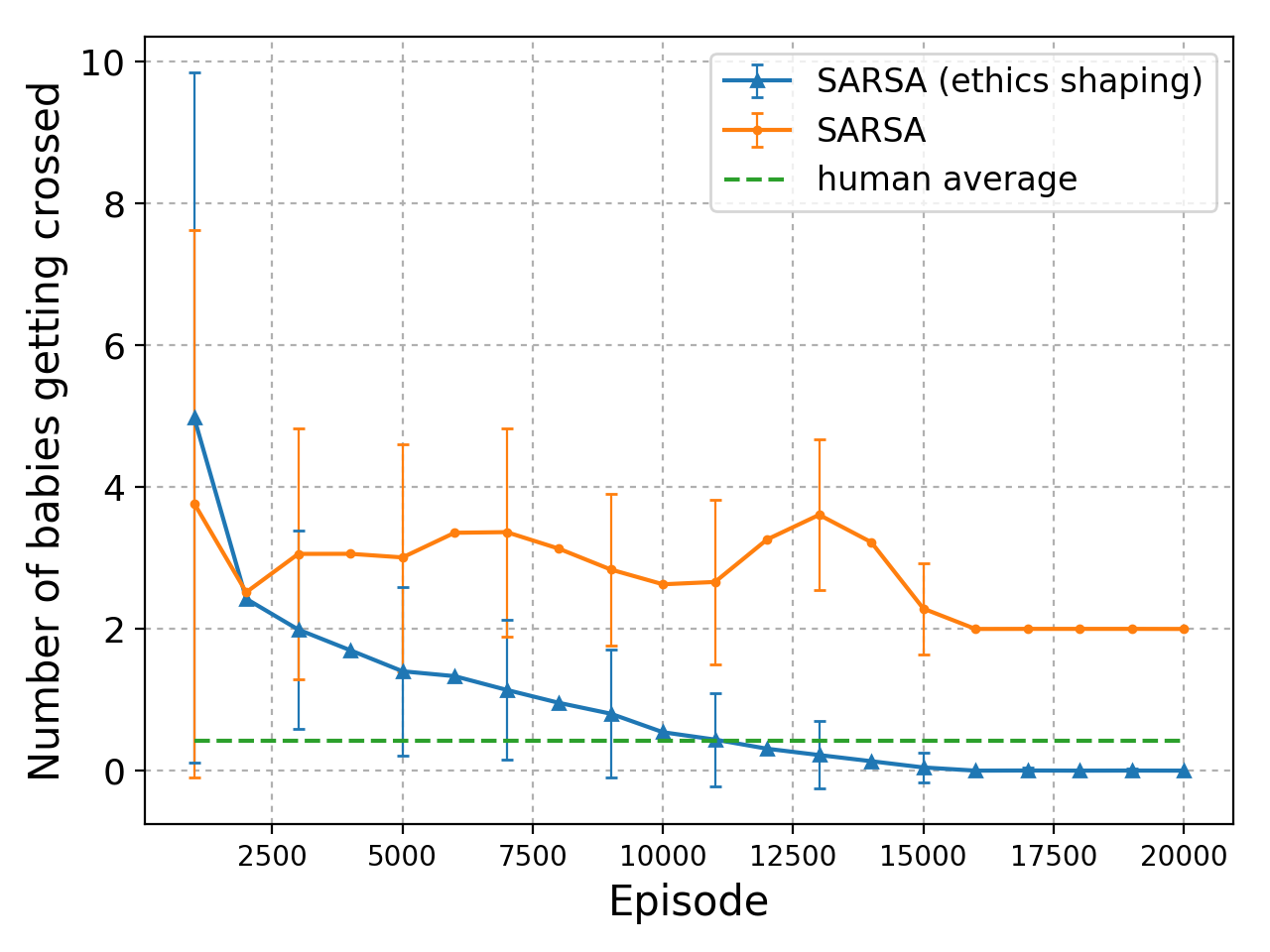}
    \caption{Number of babies crossed vs. number of episodes. Average over 1000 runs are plotted with 1 s.e. errorbars.}
    \label{fig:mobj}
\end{figure}

\begin{figure}[!ht]
    \centering
    \includegraphics[scale=0.45]{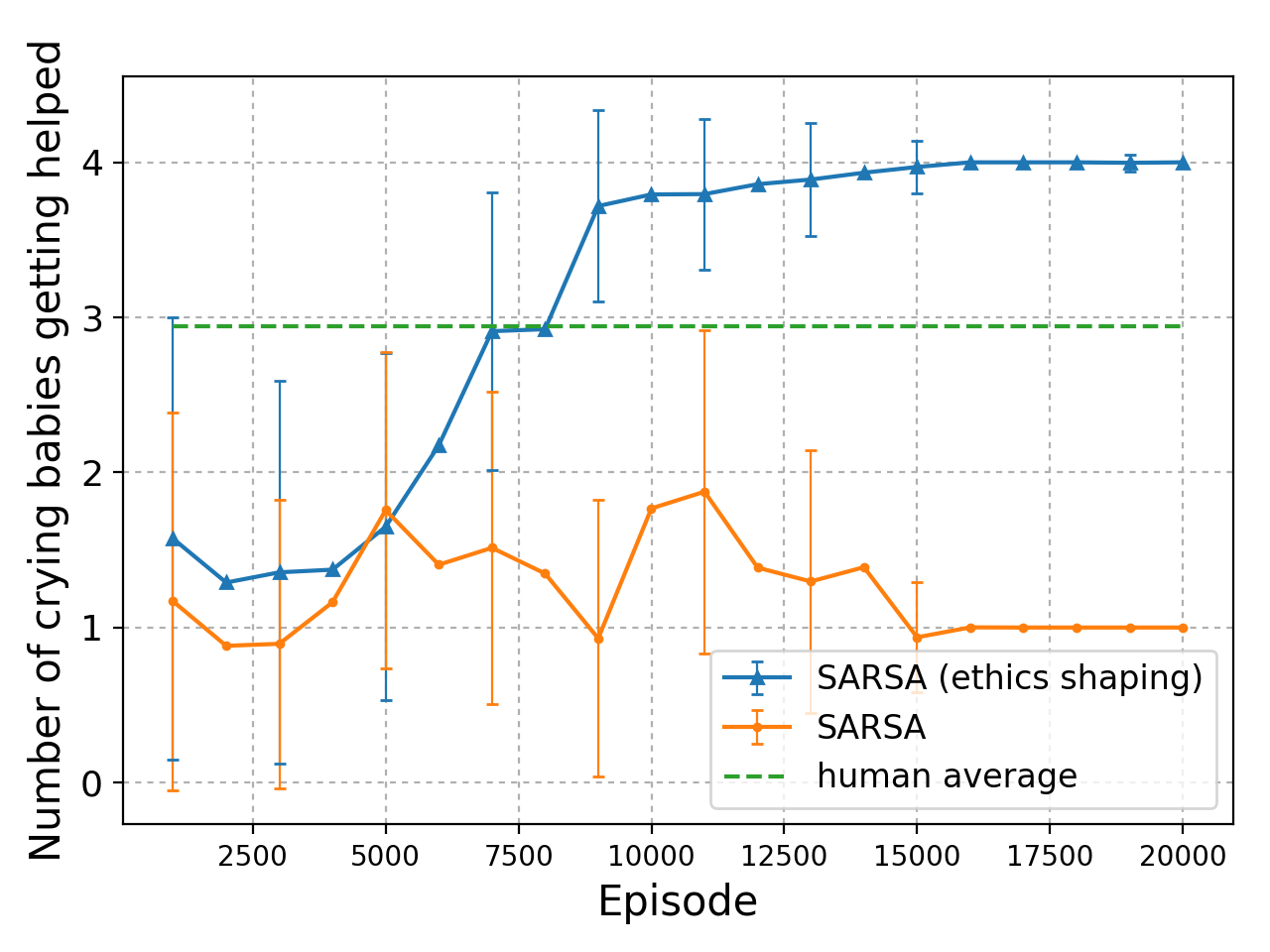}
    \caption{Number of babies getting helped vs. number of episodes. Average over 1000 runs  with 1 s.e. errorbars.}
    \label{fig:mcry}
\end{figure}

There are 48,620 optimal solutions (18 steps to the milk) for the defective reward function and  ideally there exist some routes that both avoid non-crying babies and comfort crying babies as many as possible. Figure \ref{fig:mobj} and \ref{fig:mcry} display how the agent improves at the two sub-goals through ethics shaping. Note that the agent actually helps more babies than human beings because the reward propagation mechanism in reinforcement learning makes the learner come up with more thorough plans. Unlike apprenticeship learning which directly imitates human behaviors, ethics shaping enables reinforcement learner not to be biased by inadequate decisions of human beings. Additionally, Figure \ref{fig:mperf} suggests that the extra tasks do not significantly affect the convergence.


\subsection{Driving and Avoiding}
Since autonomous cars have attracted attention for ideally being able to dramatically reduce the number of traffic accidents, some ethical issues \cite{av1,av2} have been claimed for security. We would like to deploy this toy example to demonstrate that ethics shaping is capable of dealing with driving issues when the reward function is incomplete.

Our car driving simulation is similar to the second experiment in \cite{al} except that cars could be driving in all of the lanes and sometimes there are seriously wounded cats lying in certain lanes which we should avoid so as not to make them worse. We are driving faster than all of the other cars and the cats relatively approach us the fastest since they are unable to move. Even though dying cats may not directly relate to machine ethics which usually indicates human-machine interactions, we use dying cats to represent other objects such as humans injured in car accidents or elderly people with dementia. To be a good driver, it is also encouraged to drive straight when switching lanes is not needed. The problem definition without considering ethics is as follows:\\\\
\noindent {\bf Objective} (\textit{Driving and Avoiding})
\begin{align*}
    \min_{{\bf A}=\{a_1,a_2,\cdots ,a_n  \vert a_i\in \mathcal{A} \}} L({\bf A}),
\end{align*}
where
\begin{align*}
    L({\bf A})=\sum_{a_i\in \mathcal{A}} &p_1 \cdot \mathbbm{1}[\![a\in \textit{Collision}]\!] - \\&p_2 \cdot \mathbbm{1}\mathbbm[\![a = \textit{straight}]\!],
\end{align*}
$\mathcal{A}$ is all possible actions, \textit{Collision} is the set of actions that might collide with one of the cars, and \textit{straight} is the action to drive straight. $p_1$ and $p_2$ are set to $20$ and $0.5$ respectively in our experiment.

By this experiment, we would like to test whether the ethics shaping technique is capable of making reinforcement learners dodge dying cats as well as be good drivers. The goals of this task can be stated as follows.
\begin{itemize}
    \item Primary goal: avoid collisions
    \item Sub-goals: (1) drive straight, (2) dodge dying cats.
\end{itemize}
Manually generated human trajectories aim at avoiding running over dying cats and averting car collisions. Some randomness is added to give variety. The MDP has three actions, which allow the agent to steer smoothly to one of the neighboring lanes and go straight. There are five features indicating what lane the car is currently at and the other twelve features indicating the discretized distance of the closest car and the closest cat in the left, current and the right lane respectively. The incomplete reward function is defined as the negative loss function as described above.

\begin{figure}[!ht]
    \centering
    \includegraphics[scale=0.45]{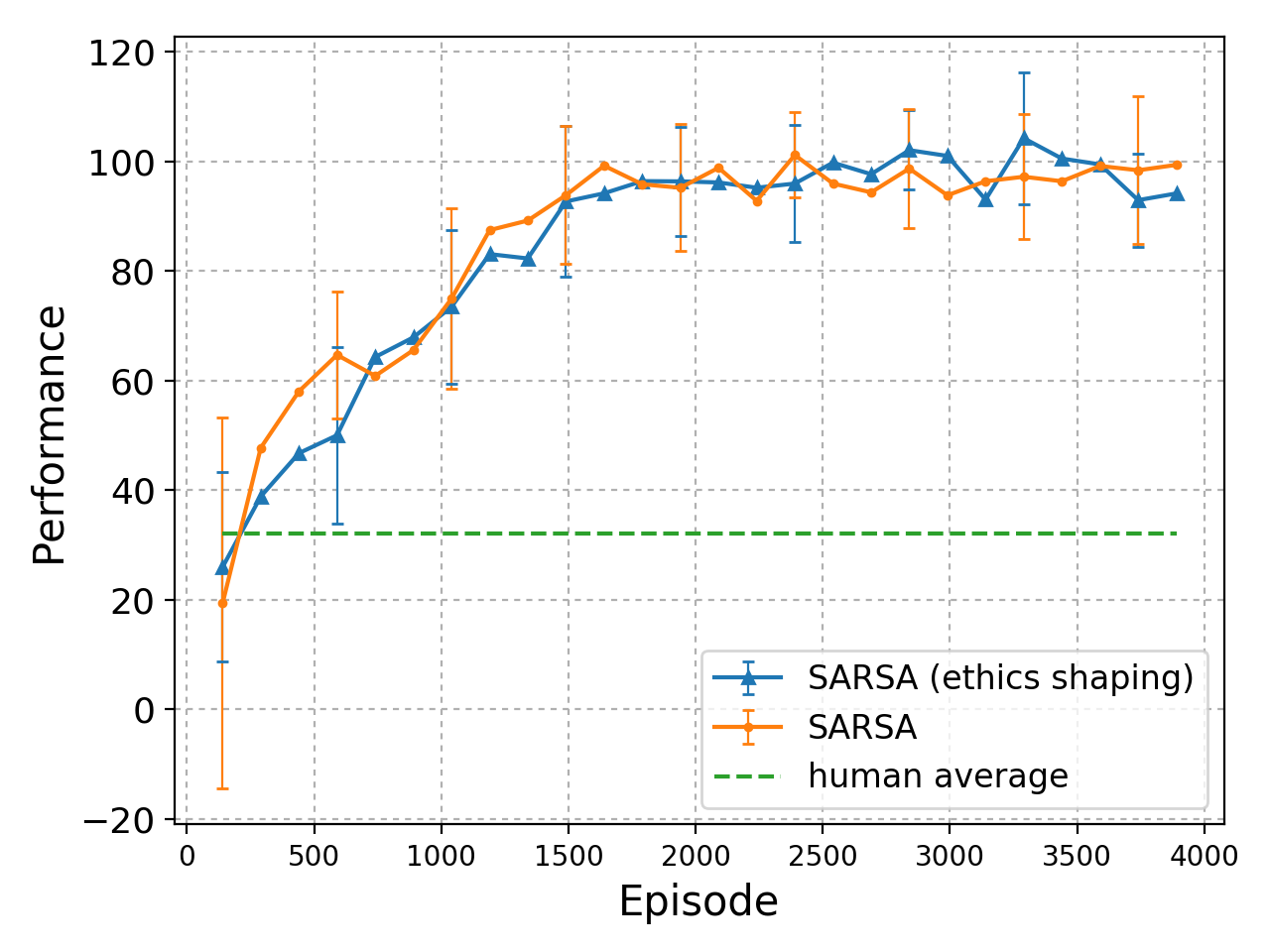}
    \caption{SARSA with and without ethics shaping in the \textit{Driving and Avoiding} experiment on cumulative rewards. Average over 150 runs  with 1 s.e. errorbars.}
    \label{fig:rreward}
\end{figure}

\begin{figure}[!ht]
    \centering
    \includegraphics[scale=0.45]{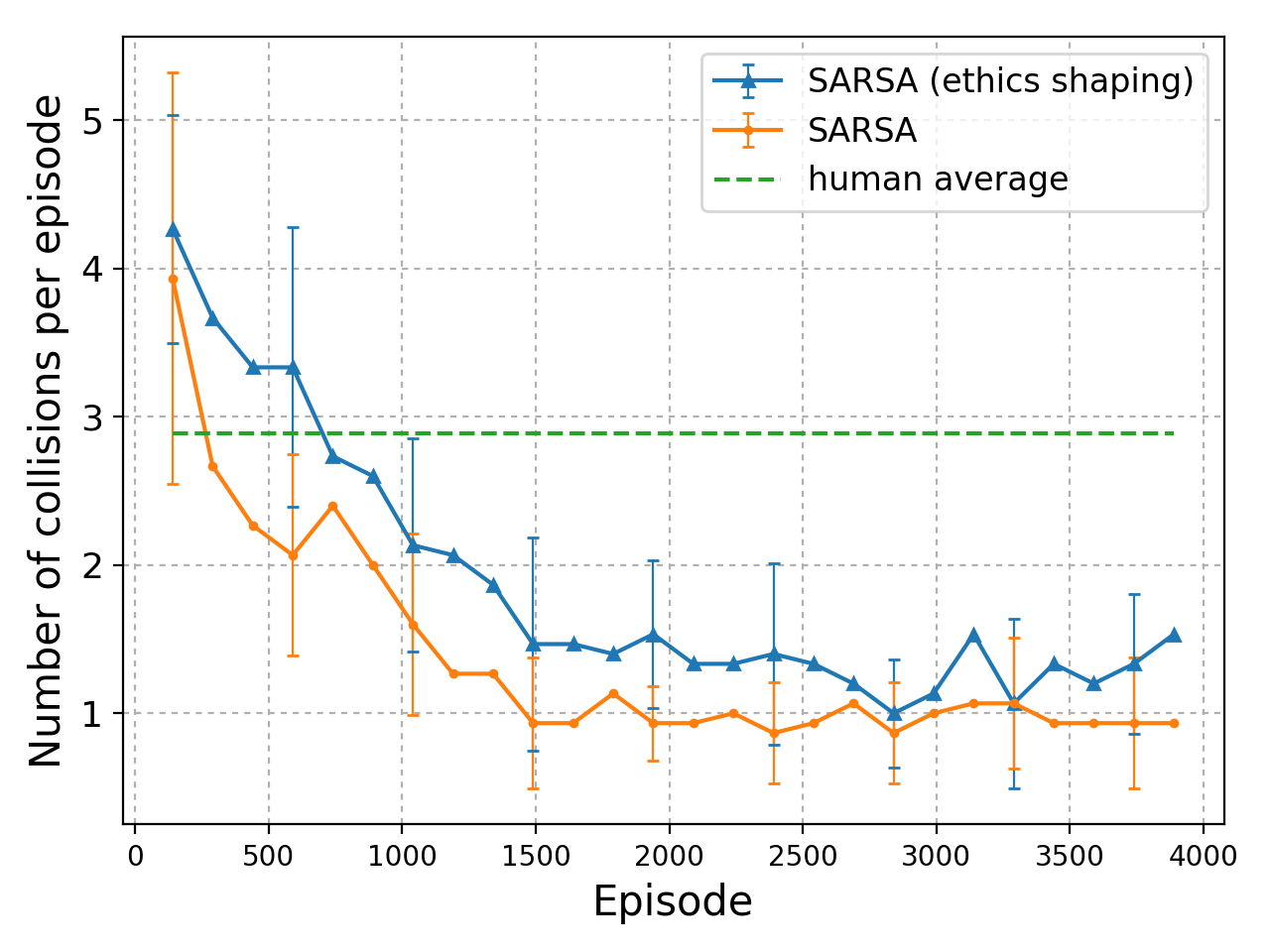}
    \caption{Number of collisions vs. number of episodes. Average over 150 runs with 1 s.e. errorbars.}
    \label{fig:rcollision}
\end{figure}

\begin{figure}[!ht]
    \centering
    \includegraphics[scale=0.45]{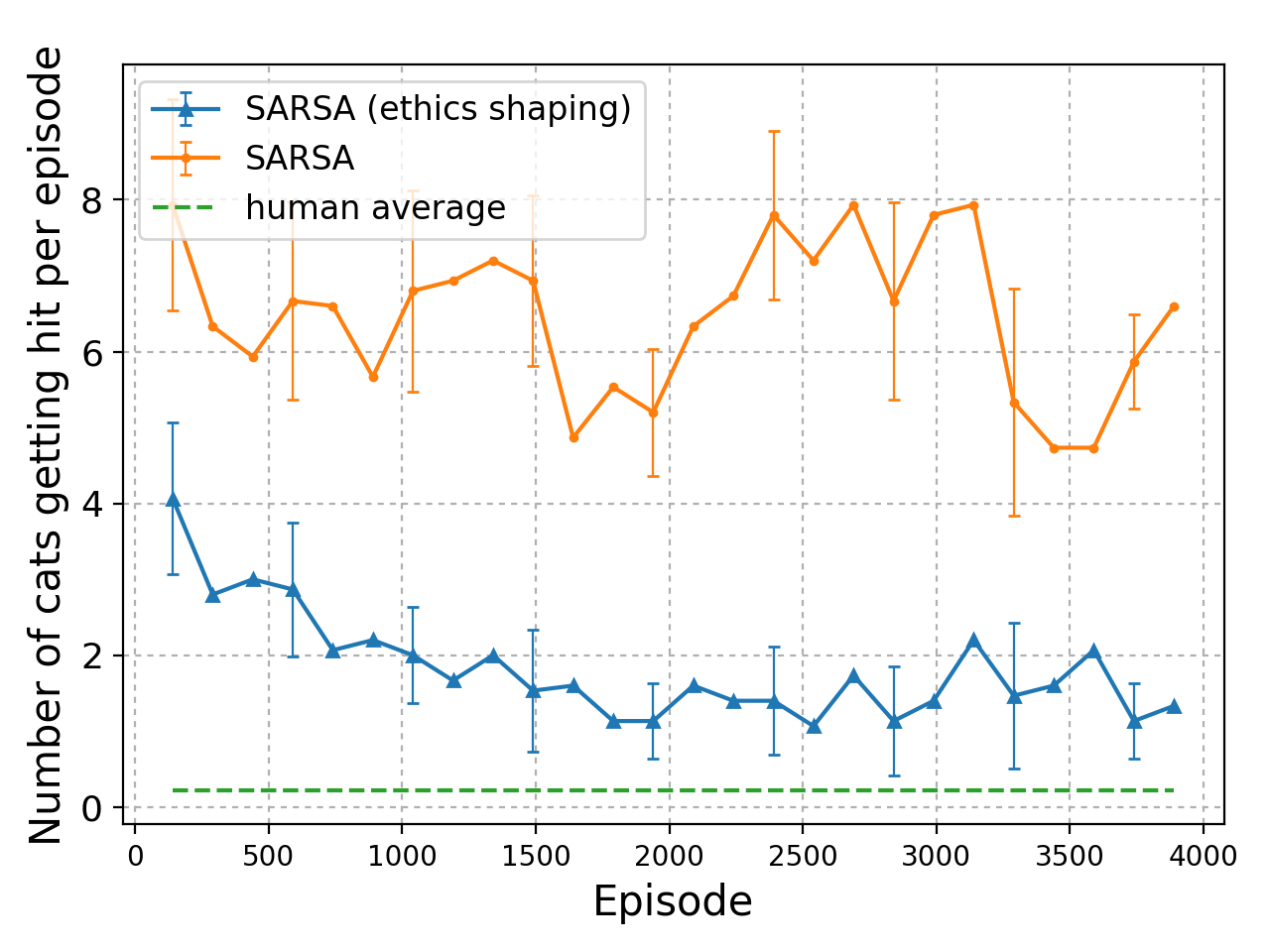}
    \caption{Number of cats getting hit within one episode. Average  over 150 runs with 1 s.e. errorbars.}
    \label{fig:rhit}
\end{figure}

This scenario is more difficult than \textit{Grab a Milk} since sometimes it is required to make decisions between collision with cars and hitting wounded cats. Collisions are occasionally unavoidable due to the limited horizon of the agent. In this experiment, the human trajectories are generated with a rule that avoiding hurting cats first and then avoiding collisions by switching to the other lanes. The performance is evaluated by cumulative reward through one episode. It is shown that ethics shaping is able to acquire descent performance and still preserve ethical behavior. As Figure \ref{fig:rreward} and \ref{fig:rcollision} suggest, in the reinforcement learning process, there is no significant difference between two algorithms with respect to cumulative rewards and number of collisions. Additionally, the significant reduction in the number of cats getting hit is shown in Figure \ref{fig:rhit}, which provides an insight that ethics shaping is able to resolve the conflicts between performances and ethical decisions. 


\subsection{Driving and Rescuing}
\textit{Driving and Rescuing} is similar to \textit{Driving and Avoiding} in terms of environments. However, in this scenario, instead of avoiding running over dying cats, the sub-task for the agent is to rescue the dementia elderly trapped in the traffic by taking them into the car. We simplify the problem by considering that to rescue the elderly it is required to drive through their positions and the process takes no time. Consequently, it is the opposite problem of \textit{Driving and Avoiding} in which the agent should avoid crossing cats. 

The problem is more challenging than \textit{Driving and Avoiding} since there are more choices to stay away from a cat; however, to rescue the elderly, the action toward them is the only option. As Figure 7 and 8 suggest, to rescue more elders, it is inevitable for human beings to experience more collisions than in \textit{Driving and Avoiding}. Even though SARSA algorithm with ethics shaping seems to perform slightly worse, it is reasonable since sacrifice (i.e. switching lanes) is needed to rescue elders. A piece of supporting evidence is that Figure 8 reveals there is no much difference between two approaches in terms of the number of collisions. With regard to the number of elders getting rescued, a significant change is shown in Figure 9, which verifies the ability of ethics shaping to make the learner behave ethically while pursuing better performance.
Another conclusion can be made in the three experiments that the problem of dangerous exploration \cite{concrete} is alleviated since in the learning process, penalties are given while the agents making unethical decisions. Consequently, the total number of unethical decision making is greatly reduced compared with original reinforcement learners.

\begin{figure}[!ht]
    \centering
    \includegraphics[scale=0.45]{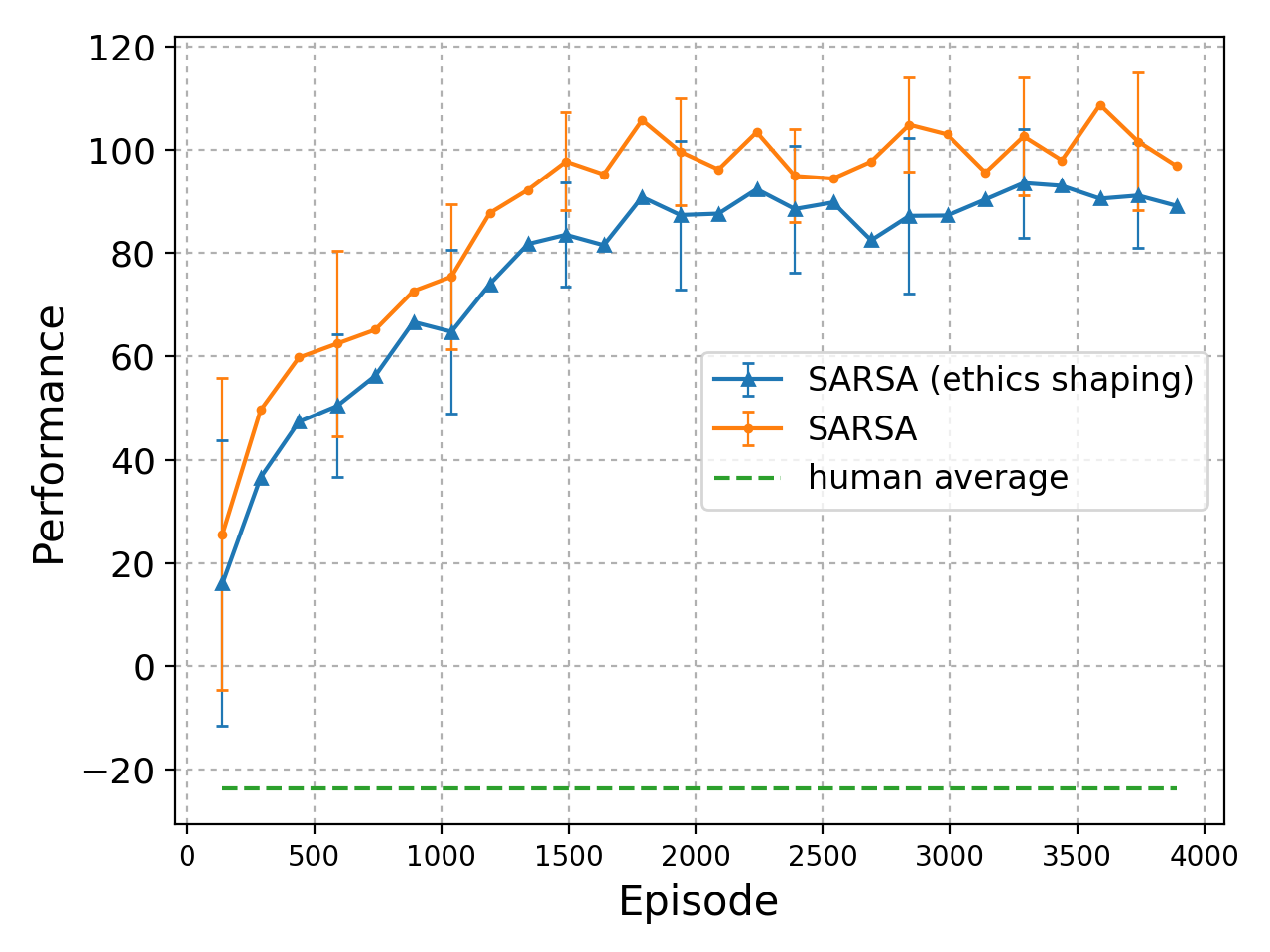}
    \caption{SARSA algorithm with and without ethics shaping in \textit{Driving and Rescuing} on cumulative rewards. Average over 150 runs are plotted with 1 s.e. errorbars.}
    \label{fig:creward}
\end{figure}

\begin{figure}[!ht]
    \centering
    \includegraphics[scale=0.45]{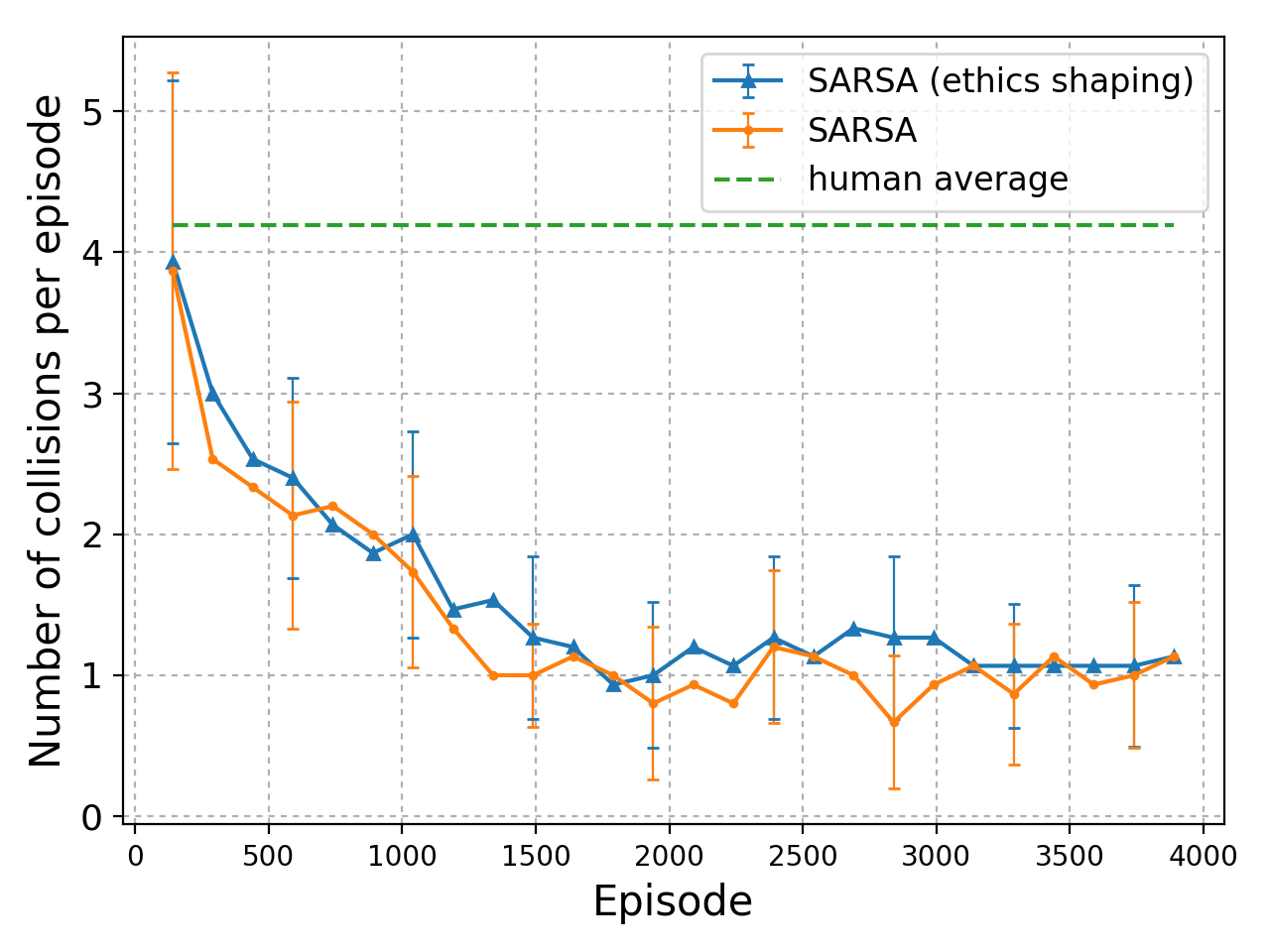}
    \caption{Number of collisions vs. number of episodes. Average over 150 runs, with 1 s.e. errorbars.}
    \label{fig:ccollision}
\end{figure}
\begin{figure}[!ht]
    \centering
    \includegraphics[scale=0.45]{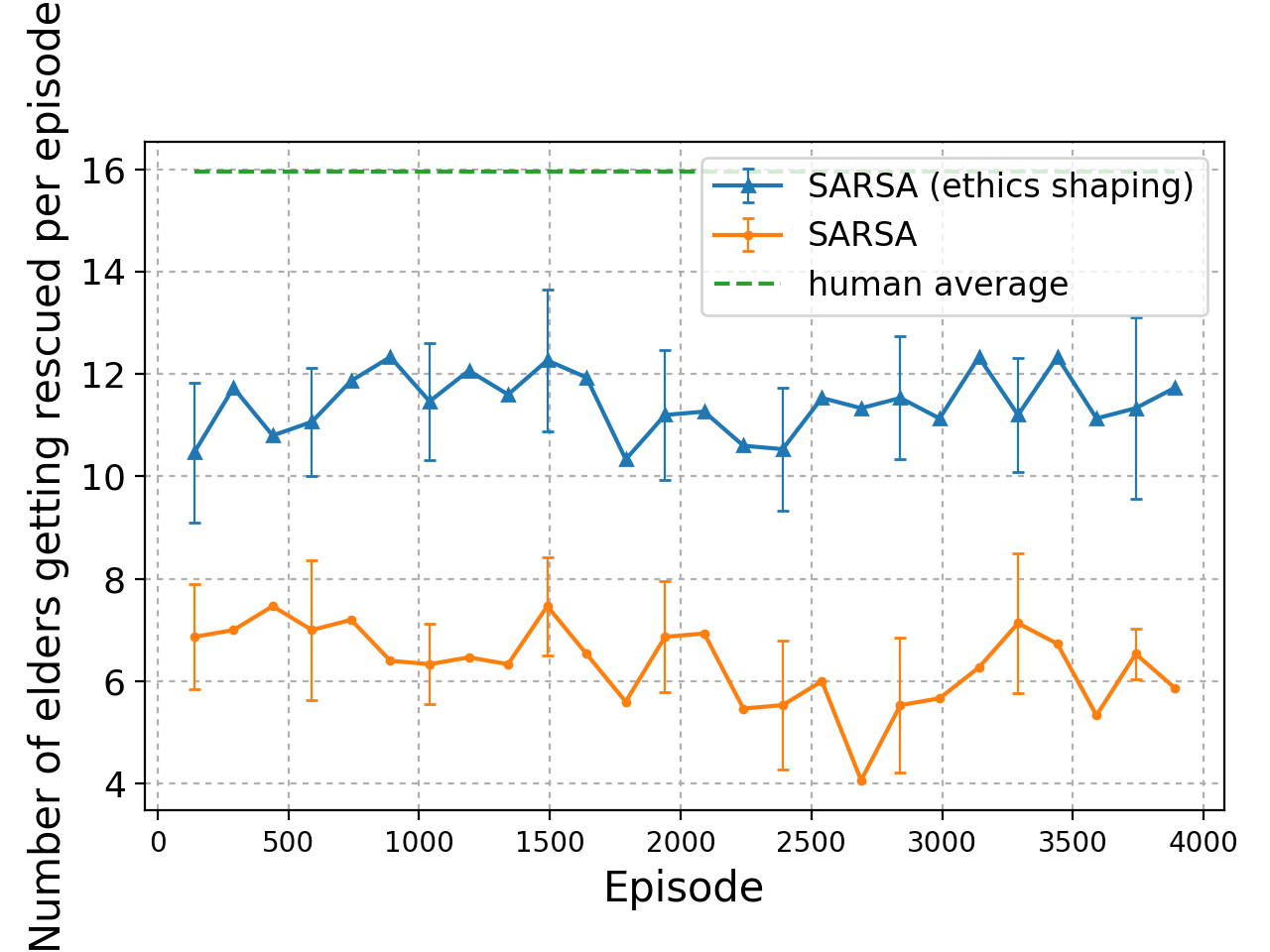}
    \caption{Number of elders getting rescued within one episode. Average over 150 runs  with 1 s.e. errorbars.}
    \label{fig:chelp}
\end{figure}

\section{Related Work}
Machine ethics \cite{me}, a project that aims to make an AI system's decision-making procedure obey some norms and ethics, has drawn attention since the AI systems have become part of the lives of modern people. Some proactive issues \cite{wirehead,bostrom,super} have been proposed to discuss possible situations that might harm the interactions between human and machines. Several issues are resulted from ill-designed objective functions \cite{concrete}, which our work aims to solve. To the best of our knowledge, the idea that employs ordinary human data to learn ethical behaviors has not been proposed. We provide a brief survey of existing approaches that relate to ethical decision making and learning. 

\subsection{Rule-Based Approaches}
\cite{sorry} proposes a mechanism to determine when and how it is best to reject directives from human interlocutors. Under their architecture, `fecility conditions' are reasoned to ensure matters such as the agent know how to accomplish the task and accomplishing the task does not violate normative principles. Those conditions are formulated as a logical expression along with inference rules. 

Horty logic \cite{horty} is a deontic logic \cite{clarke} that allows reasoning about multiple agents and their actions. \cite{deontic1,deontic2} propose similar approaches that utilize Horty logic to compose ethical semantics. However, this formalism suffers from similar limitations as Briggs and Scheutz's approach: ethical uncertainty is not allowed for decision making, active learning of the ethical rules is not permitted, and all rules should be rendered in advance. With the aid of ethics shaping, there is no need to enumerate all possible ethical rules since the integrated policy from human data is able to suggest ethical moves with our claim that most of the people would obey ethical code.

\subsection{Learning-Based Approaches}
Richer kinds of materials have been explored to achieve value alignment \cite{priority}, which is a property of an agent indicating that it can only pursue goals beneficial to humans \cite{priority,aligning}. \cite{stories} claims that stories are necessarily reflections of the culture and society; consequently, stories are a wealth of data where cultural values tacitly hold. They first generate a plot graph from crowdsourced stories using the technique described by \cite{crowdsourced}. However, stories may not be detailed enough to describe sophisticated behavior such as driving cars.


It is a challenging problem for agents to derive their objective functions while making decisions. \cite{armstrong} uses Bayesian learning to update beliefs about the utility functions that best match ethical behaviors. 
Adopting the concept of utility functions as well, \cite{rledm} considers the problem of ethical learning as learning an ethical utility function that is a part of hidden state of \textit{Partially Observable Markov Decision Process} (POMDP). The difference with Armstrong's work is that the agent is not maximizing a changing meta-utility function. Instead, the uncertainty of the ethical utility function is coupled with the uncertainty in the rest of the world. 
\cite{accountable} claims that IRL by itself is insufficient for agents to infer norms that are temporally complex, unless each state contains sufficient information to characterize the history of the agent with respect to norms. 
To combine the strength of RL and logical representations, they propose a hybrid approach that agent would prioritize adherence. The agents would maximize the reward function over only those state-action pairs that maximally satisfy the norms. 

\section{Conclusion}
Ethics shaping is proposed to make reinforcement learners not only achieve the expected performance and the goals but also comply with ethical rules. It utilizes reward shaping and stochastic policy from human data to balance ethical behavior and performance pursuit by providing additional reward. The reward is given if the move is related to ethics identified by integrated human policy. It can be incorporated with a variety of reinforcement learning algorithms since most of the reinforcement learning frameworks rely on reward functions. 

We coin three scenarios \textit{Grab a Milk}, \textit{Driving and Avoiding}, and \textit{Driving and Rescuing} to simulate real-life matters that everybody would possibly experience. In the three experiments, we show the capability of ethics shaping that it could outperform human policies with respect to positive ethical decisions (e.g., saving people) since reinforcement learners provide thorough plans even only local information is given. Additionally, although under more constraints than original problems, ethics shaping still achieves competitive performances with RL algorithms without ethics shaping.

\balance
\bibliography{reference}
\bibliographystyle{aaai}
\end{document}